\documentclass{article}
\PassOptionsToPackage{authoryear}{natbib}
\usepackage[preprint]{neurips_2023}
\usepackage[utf8]{inputenc} % allow utf-8 input
\usepackage[T1]{fontenc}    % use 8-bit T1 fonts
\usepackage{hyperref}       % hyperlinks
\usepackage{url}            % simple URL typesetting
\usepackage{booktabs}       % professional-quality tables
\usepackage{amsfonts}       % blackboard math symbols
\usepackage{nicefrac}       % compact symbols for 1/2, etc.
\usepackage{microtype}      % microtypography
\usepackage[font=small,labelfont=bf,tableposition=top]{caption}
\usepackage{amsmath,amssymb,amsfonts}
\usepackage{mathtools}
\usepackage{algorithmic}
\usepackage{graphicx}
\usepackage{subfigure}
\usepackage{textcomp}
\usepackage{xcolor}
\usepackage{array}
\usepackage{tikz}
\usetikzlibrary{positioning}

\makeatletter
	{\end{proof}%
	\renewcommand{\qedsymbol}{\qedsymbol}}
\makeatother

\title{An Encoder-Decoder Approach for Packing Circles}

\author{{Akshay Kiran Jose, Gangadhar Karevvanavar, Rajshekhar V Bhat}\\
 Indian Institute of Technology Dharwad, Dharwad, Karnataka, India\\ 
  \texttt{\{190020003, 212021007, rajshekhar.bhat\}@iitdh.ac.in} 
}

\begin{document}

\maketitle

\begin{abstract}
    The problem of packing smaller objects within a larger object  has been of interest since decades.
    In these problems, in addition to the requirement that the smaller objects must lie completely inside the larger objects, they are expected to not overlap or have minimum overlap  with each other. Due to this, the problem of packing  turns out to be a non-convex problem, obtaining whose optimal solution is challenging. As such, several heuristic approaches have been used for obtaining sub-optimal solutions in general, and provably optimal solutions for some special instances. 
    In this paper, we propose a novel encoder-decoder   architecture consisting of an encoder block, a perturbation block and a decoder block, for packing identical  circles within a larger circle.  In our approach, the encoder   takes the index of a circle to be packed as an input and outputs its center  through a normalization layer, the perturbation layer adds controlled perturbations to the center, ensuring that it does not deviate beyond the radius of the smaller circle to be packed, and the decoder   takes   the perturbed center as input and estimates the index of the intended circle for packing. We parameterize the encoder and decoder by a neural network and optimize it to reduce an error between the decoder's estimated index and the actual index of the circle provided as input to the encoder.
The proposed approach can be generalized to pack objects of higher dimensions and different shapes by carefully choosing normalization and perturbation layers. 
The approach gives a sub-optimal solution and is able to pack smaller objects  within a larger object  with competitive performance with respect to classical methods.
\end{abstract}

\section{Introduction and Motivation}
Packing objects  inside a larger object has been a topic of interest to researchers for decades due to its applications in various fields including computer graphics, electronics design, wireless communications, manufacturing, logistics, and transportation. In this work, we focus on circle packing, a common problem in which circles are arranged in a container without overlapping, or with minimum overlap if such an arrangement is not feasible. 
The circle packing problems are non-convex. To see this, consider the problem of packing $N\geq 1$ circles of radius $r$ within a circle of radius $R\geq r$, by finding the centers of the smaller circles to minimize the total overlap among the circles. This can be formulated as the following optimization problem:
\begin{subequations}\label{eq:main}
\begin{align}
&\underset{\boldsymbol{c}_i\in \mathbb{R}^n,   \epsilon_{ij}}{\textbf{minimize}}  &&O = \sum_{\substack{1\leq i<j\leq N}}\mathrm{overlap}(\epsilon_{ij}),  \\
&\textbf{subject to} \;&&\lvert\lvert \mathbf{c}_i\rVert_2 - (R - r)\leq 0,\label{ins} \\
&&&2r-\lvert\lvert \mathbf{c}_i-\mathbf{c}_j\rVert_2\leq \epsilon_{ij},\; i\neq j, \label{btw1} 
\end{align}
\end{subequations}
for all $1\leq i<j\leq N$, where the larger circle of radius $R$ is assumed to be centered at the origin $(0,0)$ and $\mathbf{c}_i$ is the center of $i^{\rm th}$ smaller circle, when circles are indexed in $\{1,2,\ldots,N\}$. 
The constraint \eqref{ins} ensures that the centers of all $N$ smaller circles of radius $r$ lie within the larger circle, and \eqref{btw1} captures restriction on the overlap between any two smaller circles. Here, $\rm overlap{(\epsilon_{ij})}$ gives the overlap area when the overlap length between $i^{\rm th}$ and $j^{\rm th}$ circles is $\epsilon_{ij}$. 
Notably, the constraint \eqref{ins} for each $i \in {1,2,\ldots,N}$ is convex, but the  constraints specified by \eqref{btw1} are non-convex. Consequently, the overall problem \eqref{eq:main} becomes non-convex, making the search for an optimal solution challenging. These non-convex constraints also arise in other variations of the circle packing problem, including cases where one seeks to determine $N$, $R$, or $r$, given values of any two of them, along with centers, under the condition of no-overlap (i.e., with $\epsilon_{ij}$ equal to zero). 

%%%%%%%%% BEST KNOWN PACKING %%%%%%%%%%%%%%%
%http://hydra.nat.uni-magdeburg.de/packing/cci/
%http://www.inf.u-szeged.hu/~pszabo/Pub/45survey.pdf

\section{Related Literature}
The circle packing problem with no-overlap condition  has been extensively studied in literature. Various techniques have been used to solve the problem, including non-linear programming, stochastic search, heuristics with local search and a neural gas method, as described below. 

\paragraph{Non-Linear Programming} 
In the work by \cite{DensePackingGraham1996}, the circle packing problem is formulated as placing $N$ points within a unit circle to maximize an approximate minimum pairwise distance between them. The authors solved this problem using a steepest descent search with Goldstein-Armijo backtracking line search and a modified Newton method.  \cite{he2020stochastic}   proposes a stochastic item descent method (SIDM)  inspired by mini-batch stochastic gradient descent for solving the equal circle packing problem. Disciplined Convex-Concave Programming (DCCP) is a framework developed to solve convex-concave programs, which includes various circle packing problems \cite{shen2016disciplined}.   In this work, we adopt the DCCP-based solution as the benchmark for the solution obtained using the encoder-decoder method. 
\paragraph{Stochastic Search Methods} 
A method called the ``Billiards Simulation''  proposed by \cite{DensePackingGraham1996},  involves initially scattering $N$ points in a unit circle and then drawing circles of suitable diameter $d$ around each point. The circles are then allowed to move chaotically while adhering to the constraints of not overlapping and not escaping outside the larger circle. The diameter $d$ gradually increases until the circles settle into position. Akiyama et al. in \cite{AkiyamaMaxiMin2003} improve on this approach using a more greedy perturbation-based method. These methods are also used for packing circles in squares in \cite{ImprovingDense2000} and \cite{CircleInSquareBook2007}. Our proposed encoder-decoder approach includes a perturbation step, similar to the above classical stochastic search methods. However, our approach utilizes an encoder-decoder architecture and employs iterative adjustments of circle positions through dynamically varying random perturbations. 

\paragraph{Heuristics}
Heuristic algorithms have also gained attention for circle packing problems, which can be  classified into construction and optimization algorithms \cite{he2017efficient}. Construction algorithms focus on placing circles one at a time while ensuring feasibility. On the other hand, optimization algorithms start with an initial solution and iteratively improve it \cite{QuasiPhysical2010, HybridSimulatedAnnealing2005, GrossoMBH2008, IteratedTabu2016}. Additionally, in \cite{NeuralGas}, a neural gas-based method is employed to pack circles by clustering random points within a larger circle and utilizing the resulting cluster centers as centers for the packed circles. However, our approach differs   as we employ an encoder-decoder approach, whereas \cite{NeuralGas} utilizes a clustering method using neural networks. 
\vspace{\baselineskip}

\begin{table}[t]
    \centering
    \caption{A summary of related literature}
    \vspace{1em}
        \label{tab:lit}
    \begin{tabular}{l p{7cm} l }
        \toprule
        Method  & Comments on Performance\\
        \midrule
        Non-Linear Programming &  Introduced packings for up to $N=65$. \\
        Stochastic Search & Improved packings for $N = 70, 73, 75, 77, 78, 79, 80.$ \\
        Construction Heuristic  & Improved packings for $66 \leq N \leq 100.$ \\
        Optimization Heuristic  & Obtained $63$ better packings for $1 \leq N \leq 200.$ Found 66 better layouts for $1 \leq N \leq 320.$ \\
          Neural Gas Method  &  A method to pack circles by clustering random points within a larger circle \\
        \bottomrule
    \end{tabular}
\end{table}

Table~\ref{tab:lit} presents a summary of important methods for circle packing.  Existing methods aim to improve upon previous solutions and handle larger instances, but achieving universal global optimization remains a challenge. The website \cite{packomania} maintains a collection of best-known solutions; however, a very few solutions have been proven to be optimal, emphasizing the inherent difficulty of the problem. Remarkably, the proof of the densest packing for $8$ and $24$-dimensional spheres earned the Fields Medal in 2022.

In recent years, neural networks have been employed to tackle challenging problems, including combinatorial and black-box optimization problems \cite{bello2017neural,alkhouri2022differentiable,Neural-BO}. Motivated by this, we propose a novel encoder-decoder approach that utilizes neural networks to solve the circle packing problem, incorporating a carefully designed normalization layer and perturbation layer. We also demonstrate that our approach can be extended to higher dimensions and different shapes.

\section{Proposed Encoder-Decoder Approach for Packing Circles in a Circle}\label{sec:CinC}
In this section, we propose our encoder-decoder approach for packing $N\geq 1$ circles of radius $r$ within a circle of radius $R\geq r$. 
\subsection{Abstract Description}

We solve \eqref{eq:main} using an architecture comprising of  an encoder, a perturbation layer and a decoder.  The encoder takes the index of a circle to be packed as the input and outputs the coordinates at which the circle must be centered at. This coordinate is then perturbed using a controlled perturbation so that the coordinate after perturbation lies within the circle of radius $r$. The decoder takes the perturbed coordinate as the input and outputs an estimate of the index of the circle. 
This is concretely described in the below.

Let the circles to be packed be indexed by $i\in\{1,2,\ldots,N\}$, $\mathbf{e}_i \in \{0,1\}^N$ be the one-hot  vector with the entry $1$ in location $i$ and $0$ everywhere else for indicating  $i^{\rm th}$ circle,  and let $f_{\theta}(\cdot)$ and $g_{\phi}(\cdot)$  be the encoder and the decoder, parameterized by $\theta$ and $\phi$, respectively. Let $\mathbf{b}_i\in \mathbb{R}^n$ be the output of the encoder at the penultimate layer with $n=2$ nodes, which is normalized to obtain the center, $\mathbf{c}_i\in \mathbb{R}^n$ such that $\lvert\lvert \mathbf{c}_i\rVert_2 \leq R-r$, thereby ensuring satisfaction of constraint \eqref{ins}. This is input to the perturbation layer, where $\mathbf{c}_i$ is perturbed by adding a continuous random vector, $\mathbf{W}$ with support being a circle centered at $(0,0)$ and radius $r$.  
The perturbation layer outputs, $\tilde{\mathbf{c}}_i =  \mathbf{c}_i + \mathbf{W}$, which is given as an input to the decoder. The decoder outputs  an estimate of the index of the input circle, $\hat{\mathbf{e}} = g_{\phi}(f_{\theta}(\mathbf{e}_i)+\mathbf{W})$.

We aim to minimize the loss $({1}/{N})\sum_{i=1}^N\mathbb{E}_{\mathbf{W}}[\mathbb{I}_{\hat{i} \neq i}]$, where $\mathbb{I}$ is an indicator variable that equals $1$ if $\hat{i} = i$ and $0$ otherwise, and $\hat{i}$ is the index of the location of the maximum value of $\hat{\mathbf{e}}$. This obtains a circle packing solution that avoids overlaps whenever possible, and minimizes their extend in cases where overlaps are unavoidable. 
However, in practice, limitations on possible choices of encoder and decoder functions, not being able to sample uncountably infinite values from $\mathbf{W}$, and non-differentiability of the loss function, hinder the ability to achieve the desired result. Therefore, we present a practical compromise that yields good performance, which we describe below.

\subsection{Implementation Details}
Below, we describe the hypothesis classes for the encoder and decoder functions, the selection of the perturbation random variable, the loss function used, and the optimization method employed to achieve circle packing using the above encoder-decoder approach.
\subsubsection{Neural Network Architecture}
\begin{figure}[t]
    \centering
    \begin{center}
\begin{tikzpicture}[scale=0.6,xscale = 0.9, transform shape]
\tikzstyle{encoder}=[draw, minimum height=3cm, fill=blue!20, text width=1.5cm,align=center]
\tikzstyle{decoder}=[draw, minimum height=3cm,fill=red!20, text width=1.5cm,align=center]
\tikzstyle{bottleneck}=[draw, minimum height=2cm, fill=gray!20, text width=2.5cm,align=center]

% Input

% Encoder Layers
\node[encoder] (layer1) at (3,0) {$N$, SeLU};
\node[encoder] (layer2) at (5.5,0) {$N$, SeLU};
\node[bottleneck,fill=blue!20] (layer3) at (8.5,0) {$n$, tanh};
\node[bottleneck,fill=blue!20] (layer4) at (12,0) {$n$,   Normalization};
\node[bottleneck] (layer5) at (15.5,0)  {$n$,  Perturbation};
\node[decoder] (layer6) at (18.5,0) {$N$, ReLU};
\node[decoder] (layer7) at (21,0) {$N$, ReLU};
\node[decoder] (output) at (23.5,0) {$N$, SoftMax};
 
% Connections
\draw[->] (1,0) -- (layer1) node[above, midway] { $\mathbf{e}_i$};
\draw[->] (layer1) -- (layer2);
\draw[->] (layer2) -- (layer3);
\draw[->] (layer3) -- (layer4)   node[above, midway] { $\mathbf{b}_i$} ;
\draw[->] (layer4) -- (layer5)  node[above, midway] { $\mathbf{c}_i$} ;
\draw[->] (layer5) -- (layer6)  node[above, midway] { $\tilde{\mathbf{c}}_i$};
\draw[->] (layer6) -- (layer7);
\draw[->] (layer7) -- (output);
\draw[->] (output) -- (25.5,0)  node[above, midway] { $\hat{\mathbf{e}}$};

\end{tikzpicture}

    \end{center}
    \caption{The architecture of the neural network proposed for solving the circle packing problem. Our main contribution lies in conceiving the  encoder-decoder approach for this task, and in designing a customized normalization layer and perturbation layer that are critical to achieving good results. The layers shown in light blue, light gray and light red colors are encoder, perturbation and decoder layers, respectively. The number of nodes and activation functions used in each layer of the encoder and decoder blocks are mentioned. $N$ denotes the number of objects to be packed, with $n$ indicating the dimensionality of the objects. For circles, $n$ is $2$, whereas for spheres, $n$ is $3$.}
    \label{fig:NN}
\end{figure}
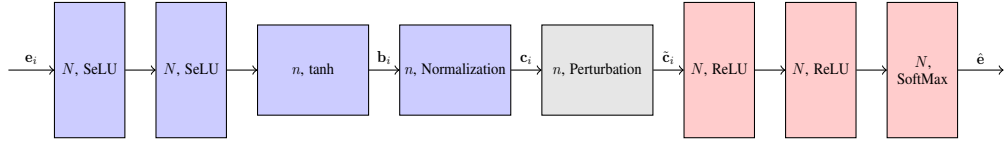

In our encoder-decoder approach, we utilize a fully connected feedforward neural network for both the encoder and decoder. An illustration of the architecture is provided in Fig.~\ref{fig:NN}. 
\paragraph{Encoder and Normalization Layer}
The encoder layer takes the one-hot vector, $\mathbf{e}_i$, as the input. 
We employ $\verb|SeLU|$ activations proposed in \cite{klambauer2017selfnormalizing} for the initial layers of the encoder.  Additionally, we use $\tanh$ as the activation function in the layer immediately preceding the normalization layer of the encoder. 
The output of this layer is $\mathbf{b}_i\in [-1,1]^n$ with $n=2$. 
The final layer of the encoder normalizes $\mathbf{b}_i$ to obtain the center $\mathbf{c}_i$, i.e., $\mathbf{c}_i = f_{\theta}(\mathbf{e}_i)$, using the following equation: $\mathbf{c}_i=\mathbf{b}_i\odot \boldsymbol{\alpha}_i$, where $\boldsymbol{\alpha}_i\in \mathbb{R}^2$ is a learnable parameter that satisfies the inequality $\lVert\boldsymbol{\alpha}_i\rVert_2\leq R-r$. %Here, $R$ and $r$ denote the radii of the larger and smaller circles, respectively. 
We enforce this inequality constraint using the \verb|clip_by_norm| function in TensorFlow \cite{tensorflow2015-whitepaper}. This constraint guarantees that $\lvert\lvert \mathbf{c}_i\rVert_2\leq R-r$, thereby ensuring that the smaller circles lie within the larger circle, as prescribed by the constraint \eqref{ins}.
The output $\mathbf{c}_i$ is perturbed to obtain $\mathbf{\tilde{c}}_i$, in the perturbation layer, described below. 
\paragraph{Perturbation Layer}
Since the smaller objects to be packed are circles, we adopt a perturbation with a random vector whose support is  a circle with center $(0,0)$ and radius equal to the radius of the circles to be packed, which can be derived as below.
\begin{itemize}
    \item Sample random variables $V_1,V_2,\ldots, V_n$ independently and identically from the standard normal distribution, i.e., $V_i\sim \mathcal{N}(0,1)$ for $i=1,2,\ldots,n$. And, sample a uniformly distributed random variable, $Z$, i.e., $Z\sim \mathcal{U}[0,1]$, independent of $V_1,V_2,\ldots, V_n$. 
    \item Derive a sample from  $\mathbf{W} = r\times Z^{p}\times\mathbf{V}/\lVert \mathbf{V}\rVert_2$, where $\mathbf{V} = [V_1~V_2~\ldots~V_n]^T$.   Here, the parameter $p\geq 0$ controls the density of sampled points at the center of the circle relative to those near the circumference. Specifically, a larger value of $p$ leads to denser points near the center, while a smaller value of $p$ results in denser points near the circumference. 
\end{itemize}
This layer adds the derived random variable, $\mathbf{W}$ to the center, $\mathbf{c}_i$ and outputs $\mathbf{\tilde{c}}_i = \mathbf{c}_i+\mathbf{W}$. 

\paragraph{Decoder Layer} The obtained $\mathbf{\tilde{c}}_i$ is passed through the layers of the decoder whose initial layers have $\verb|ReLU|$ activations. The final layer with $N$ nodes has the $\verb|SoftMax|$ activation and it outputs $\hat{\mathbf{e}}$, a probability distribution on the  indices of the circles. 
\subsubsection{Loss Function and Optimization}
Initially, the neural network weights are initialized via Xavier uniform initializer  \cite{Glorot2010UnderstandingTD}. The optimization process involves iteratively utilizing one-hot vectors representing all the circles simultaneously, at each iteration.  That is,  at each iteration, the batch $[\mathbf{e}_1~\mathbf{e}_2~ \ldots ~\mathbf{e}_N ]$ is input to the network. 
This batch is equal to the identity matrix of dimension $N$. 
The network then outputs the centers, $[\mathbf{c}_1~\mathbf{c}_2~ \ldots ~\mathbf{c}_N ]$. Then, $N$ independent samples of $\mathbf{W}$ are obtained, and added to $[\mathbf{c}_1~\mathbf{c}_2~ \ldots ~\mathbf{c}_N ]$, whose output is passed through the decoder to obtain $[\hat{\mathbf{e}}_1~\hat{\mathbf{e}}_2~ \ldots ~\hat{\mathbf{e}}_N ]$. 
We consider the following cross-entropy loss: $-\sum_{i=1}^{N}\sum_{j=1}^{N}\mathbf{e}_i^{(j)}\ln{\hat{\mathbf{e}}_i^{(j)}}$,  where $\mathbf{e}_i^{(j)}$ and $\hat{\mathbf{e}}_i^{(j)}$ are the $j^{\rm th}$ entries of vectors $\mathbf{e}_i$ and $\hat{\mathbf{e}}_i$, respectively and minimize it for a large number of iterations. We use the ADAM optimizer with learning rate of $5\times 10^{-4}$ for optimization. 
The distribution of perturbation within the circle is crucial in determining the effectiveness of our approach in achieving a satisfactory packing. To this end, we consider $p$ as a hyperparameter and employ dynamic adjustment during optimization. Through empirical observation, we find that initializing with $p=2$ and later transitioning to smaller values, such as $p=1/5$ or $p=1/10$, leads to improved packing compared to using a fixed $p$. The choice of $p$ and its impact are discussed in more detail in subsequent sections.

%we start with $p=2$ to sample denser points in the middle and quickly reduce it to $p=1/5$ within a few epochs. We maintain $p$ at this value for about $20$ epochs and then keep it as small as possible. By adapting the value of $p$ in this way, we are able to optimize the packing process and achieve better results.

\subsection{Discussion}
Our approach solves the circle packing problem by iteratively adjusting circle positions and applying perturbations, by adding a random variable with an appropriate circle as its support, to the center output by the encoder. During optimization, we adopt a higher value of $p$ in the initial phase to concentrate noise near the centers and a lower value of $p$ in the latter phase for more densely distributed noise towards the circumference, resulting in further separation,  gradually moving towards satisfying the constraint, \eqref{btw1}.   
Our proposed encoder-decoder approach can be considered a type of stochastic search method, akin to the Billiards Simulation technique proposed by \cite{DensePackingGraham1996}. However, unlike these traditional stochastic search methods, our approach employs an encoder-decoder architecture and iteratively adjusts circle positions through random perturbations while dynamically varying the perturbation distribution during optimization. This approach achieves state-of-the-art results in packing circles within circles, and it can be generalized to pack objects of other shapes and dimensions within objects of other shapes and dimensions, as described next.

\section{Extensions to Other Shapes and Higher Dimensions}\label{sec:OtherShapes}
In our encoder-decoder approach for circle packing, the normalization layer in the last layer of the encoder determines the shape of the larger object and the support of the random variable adopted in the perturbation layer determines the shape of the smaller objects. 
By increasing the number of nodes in the final layer of the encoder, $n$, from $2$ to $3$ and applying  $\lVert\cdot\rVert_2\leq R-r$ using \verb|clip_by_norm|, in the normalization layer,  we can pack spheres of radius $r$ in a larger sphere of radius $R$. 
For packing circles of radius $r$ within a square of side length $s$, we use, $n=2$ and apply $\lVert\cdot\rVert_{\infty}\leq s/2-r$, which can be implemented using using the \verb|clip_by_value| function in    TensorFlow \cite{tensorflow2015-whitepaper}.
This approach can be extended to accommodate higher dimensions and diverse shapes by adjusting $n$ and modifying the norm constraint in the final layer of the encoder, as well as adapting the distribution of the random variable in the perturbation layer. 

\section{Results and Performance Comparison with Existing Techniques}\label{sec:Comparison}

To evaluate the performance of our framework, we consider four packing scenarios: circles in circles, circles in squares, spheres in spheres, and spheres in cubes. We evaluate the effectiveness of our approach by comparing it to disciplined convex-concave programming (DCCP) using the DCCP package \cite{dccp}. We utilize DCCP to solve \eqref{eq:main}, with the total overlap length serving as the objective function proxy for compatibility with DCCP.

\begin{figure}[htp!]
\centering
\subfigure[Encoder-decoder, $7$ circles]{\includegraphics[width=0.45\linewidth]{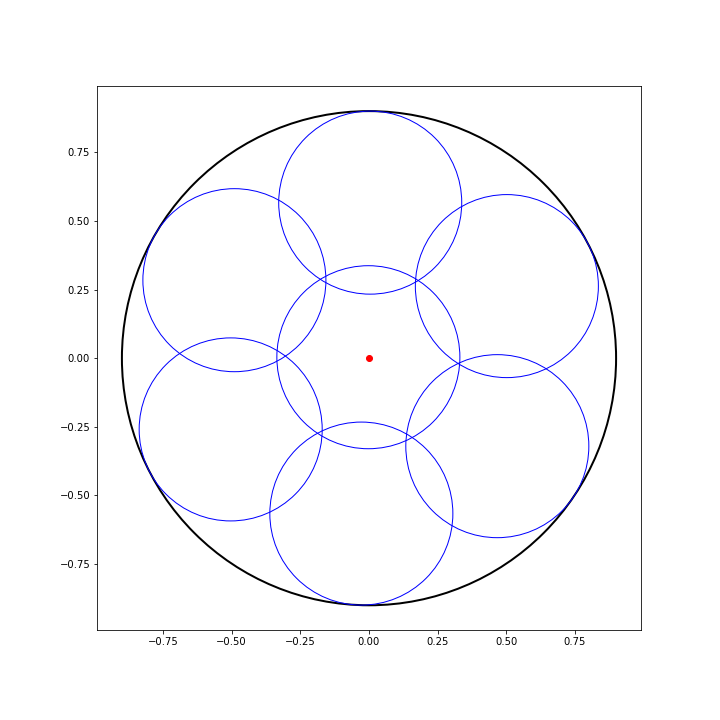}}
\hfill
\subfigure[DCCP, $7$ circles]{\includegraphics[width=0.45\linewidth]{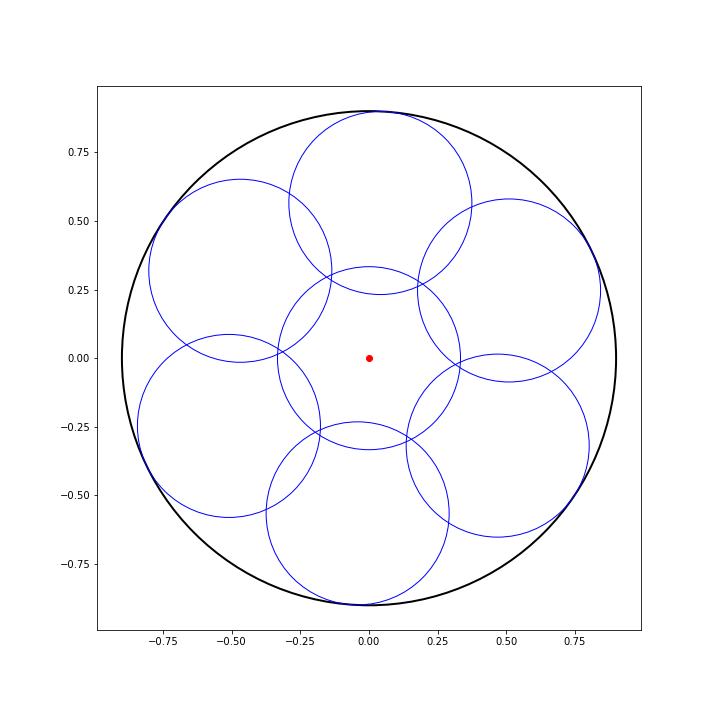}}
\bigskip
\subfigure[Encoder-decoder, $13$ circles]{\includegraphics[width=0.45\linewidth]{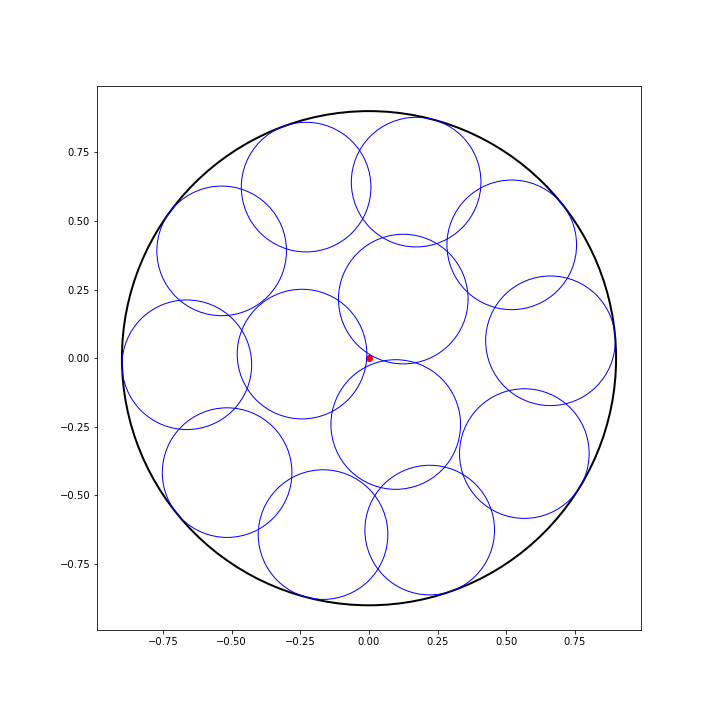}}
\hfill
\subfigure[DCCP, $13$ circles]{\includegraphics[width=0.45\linewidth]{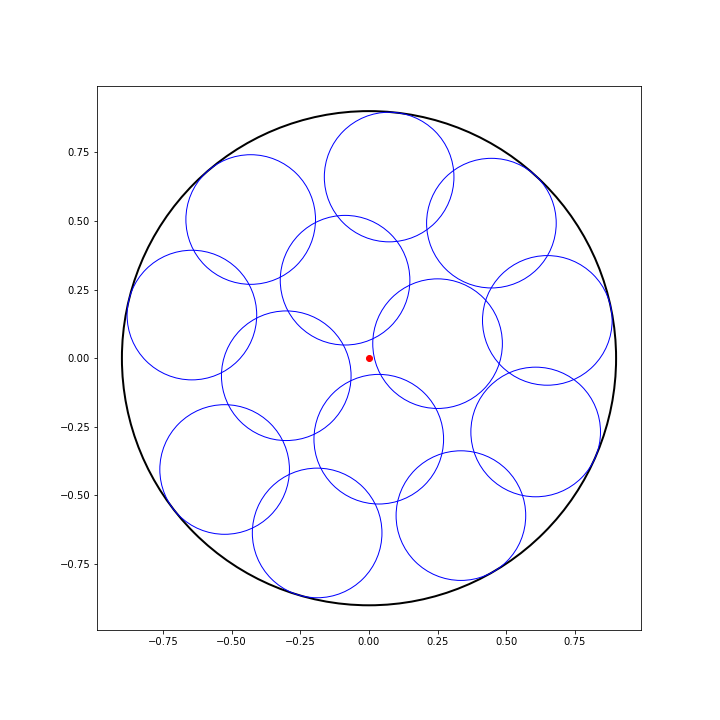}}
\bigskip
\subfigure[Encoder-decoder, $14$ circles]{\includegraphics[width=.45\linewidth]{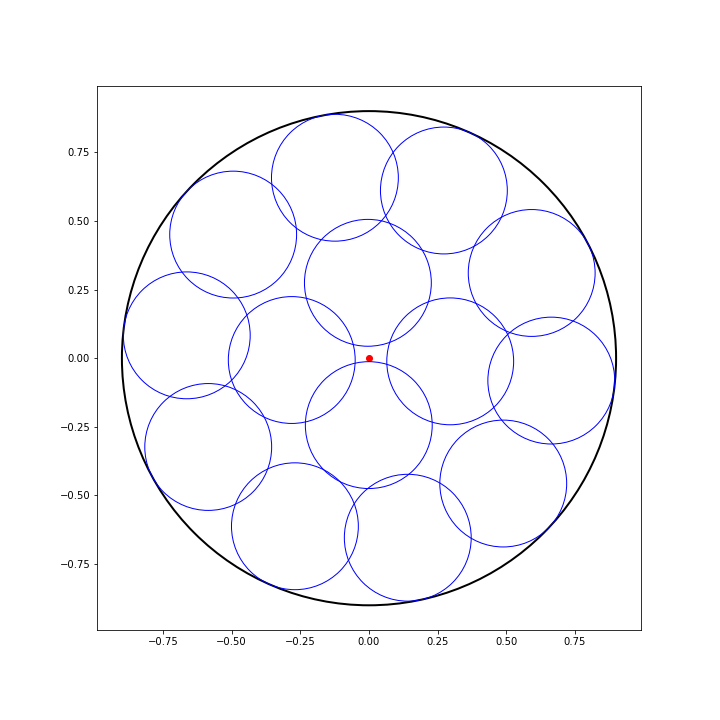}} 
\hfill
\subfigure[DCCP, $14$ circles]{\includegraphics[width=.45\linewidth]{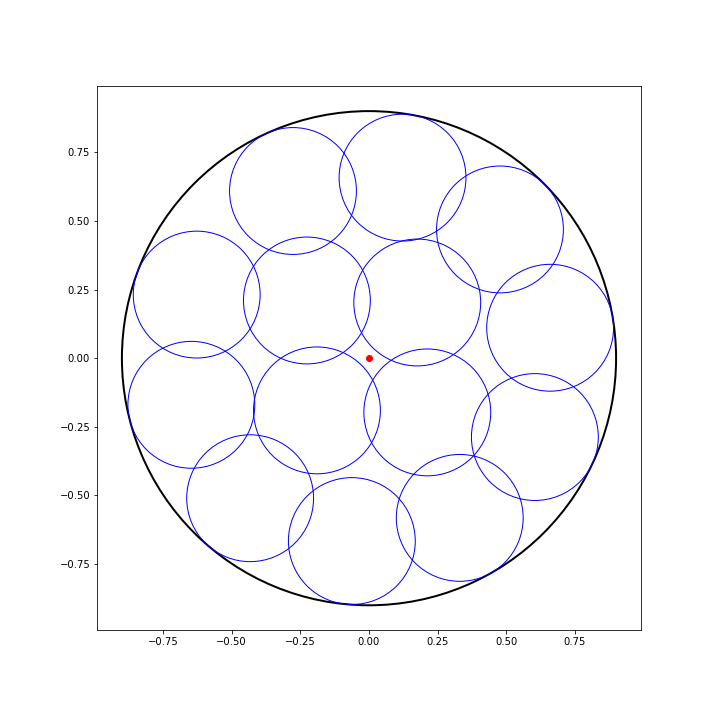}}
\caption{Arrangement of circles within a larger circle with a radius of $0.9$ units. The smaller circles have radii of $0.33333$, $0.2360679775$, and $0.2310307$ units for the cases involving $7$, $13$, and $14$ circles, respectively. Note that non-overlapping placement requires the larger circle to be of radius   at least $1$ unit. The center of the large circle is marked in red color. 
}
\label{fig:packings}
\end{figure}

Fig.~\ref{fig:packings} illustrates the packings of various circle configurations within a circle using the encoder-decoder approach and the DCCP framework. The radii of the circles are such that non-overlapping packing is not feasible. We observe that both approaches yield satisfactory packings in all scenarios. Under the DCCP framework, where we minimize the total overlap length for simplicity, all circles exhibit identical overlaps. For the packing of seven circles, both the encoder-decoder approach and DCCP produce identical results. However, for the packing of  thirteen  and fourteen circles, the resulting packings differ between the two methods.

\begin{figure}[t]
\centering
\subfigure[Spheres in a Sphere]{\includegraphics[width=0.45\linewidth]{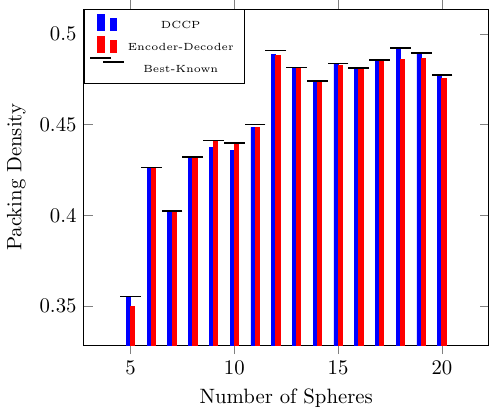}}
\hfill
\subfigure[Spheres in a Cube]{\includegraphics[width=0.45\linewidth]{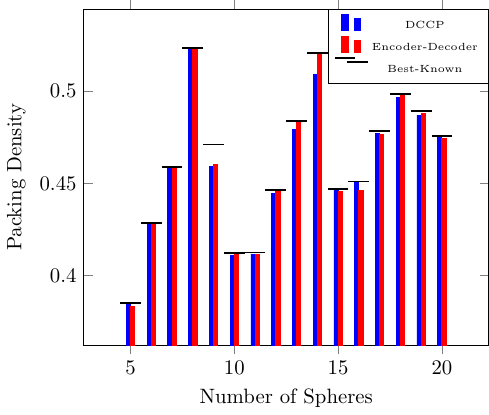}}
\bigskip
\caption{Packing Density for packing different number of spheres in a sphere and a cube.  The packing density is calculated as $((4N/3)\pi r^3 - O)/((4/3)\pi R^3)$ for the sphere case and $((4N/3)\pi r^3 - O)/s^3$ for the cube case, where $O$ represents the total overlap volume.}
\label{fig:density_dccpvsae}
\end{figure}

Fig.~\ref{fig:density_dccpvsae} displays the packing density of spheres within a larger sphere and a cube. The packing density is calculated as $((4N/3)\pi r^3 - O)/((4/3)\pi R^3)$ for the sphere case and $((4N/3)\pi r^3 - O)/s^3$ for the cube case, where $O$ represents the total overlap volume, computed along the lines in the objective function in \eqref{eq:main}. We estimate the packing density by employing Monte Carlo sampling, where a large number of points are randomly generated within the larger object. The number of overlapping points is then computed to determine the packing density.

The radii of the circles and the side length of the cube are chosen to enable non-overlapping packing. The figures present the packing density achieved by the encoder-decoder approach, DCCP, and the best known packing density sourced from \cite{packomania}. It is observed that the encoder-decoder approach performs comparably or even surpasses the performance of DCCP, and in some cases it matches the best known packing density. However, there is still potential for further improvement in packing density using encoder-decoder approach.

\begin{figure}[t]
\centering
\subfigure[When $p$ is fixed throughout optimization]{\includegraphics[width=0.45\linewidth]{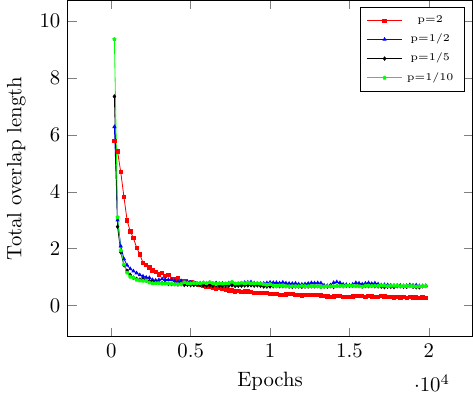}\label{fig:fixedP}}
\hfill
\subfigure[When $p$ is reduced as optimization progresses]{\includegraphics[width=0.45\linewidth]{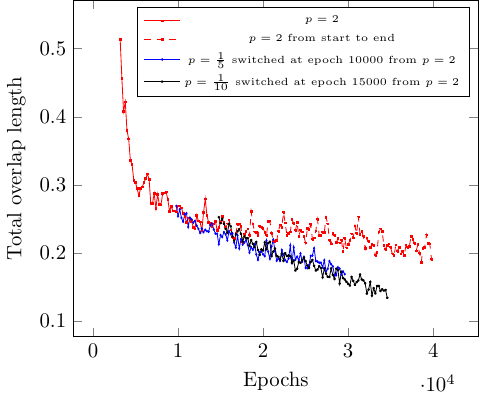}\label{fig:varyP}}
\caption{Total  overlap length versus epoch for fixed and varying values of $p$, the parameter that governs the distribution of $\mathbf{W}$, where higher values of $p$ result in a greater concentration of samples around the center of the circle.}
\label{fig:overlap_and_p}
\end{figure}

We analyze the influence of the parameter $p$ on the optimization process in Fig.~\ref{fig:overlap_and_p}. Recall that this parameter governs the distribution of $\mathbf{W}$, where higher values of $p$ result in a greater concentration of samples around the center of the circle.
We examine the effect of $p$ on the total overlap length,  $\sum_{\substack{1\leq i<j\leq N}|\mathrm{circles}~i~\mathrm{and}~j~\mathrm{overlap}} \epsilon_{ij}$,  during optimization.
From Fig.~\ref{fig:fixedP}, we observe  that setting $p=2$, which concentrates the perturbation near the center of the circle to be packed, leads to a better packing with a smaller total overlap length than when $p$ is smaller. 
The packing can be further optimized by fine-tuning the $p$ values, as shown in Fig.~\ref{fig:varyP}. From the figure, we note that starting with $p=2$ initially and subsequently switching to smaller values like $p=1/5$ or $p=1/10$ allows for further reduction in the total overlap length. This strategy ensures that the circle centers move apart sufficiently at the beginning, minimizing significant overlaps. Then, by decreasing $p$ and sampling perturbations near the circumference, the circles can continue to move and further reduce overlaps.

\begin{figure}[t]
\centering
\subfigure[An initial packing of $15$ spheres in a cube]{\includegraphics[width=0.45\linewidth]{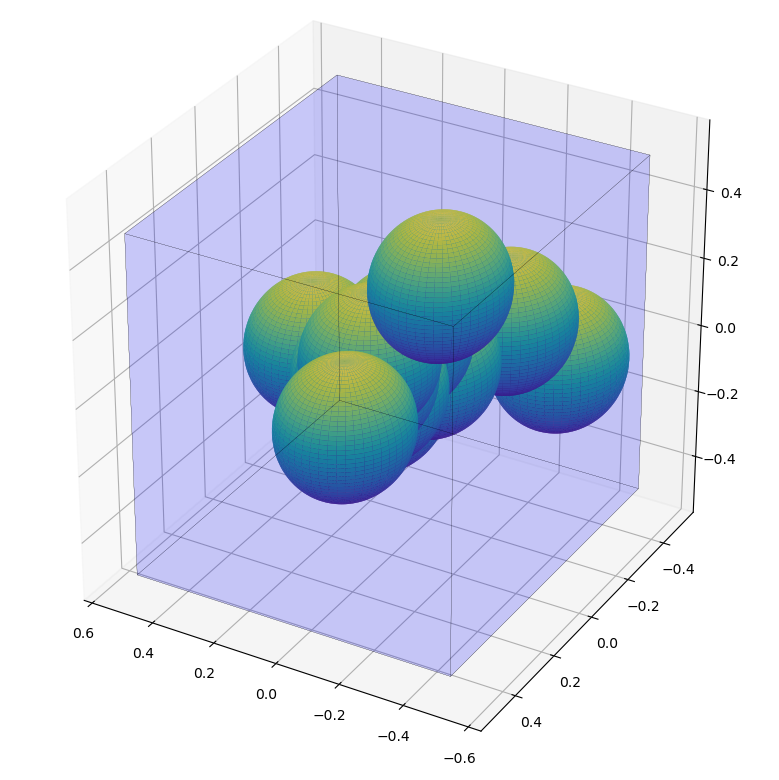}}
\hfill
\subfigure[The final packing of $15$ spheres in the cube]{\includegraphics[width=0.45\linewidth]{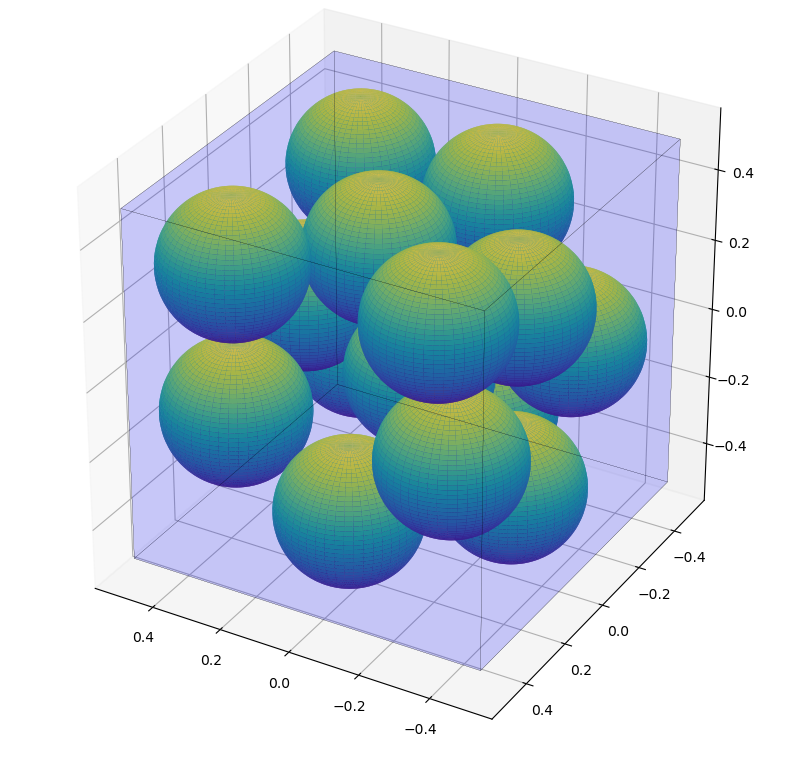}}
\bigskip
\subfigure[An initial packing of $15$ spheres in a sphere]{\includegraphics[width=0.45\linewidth]{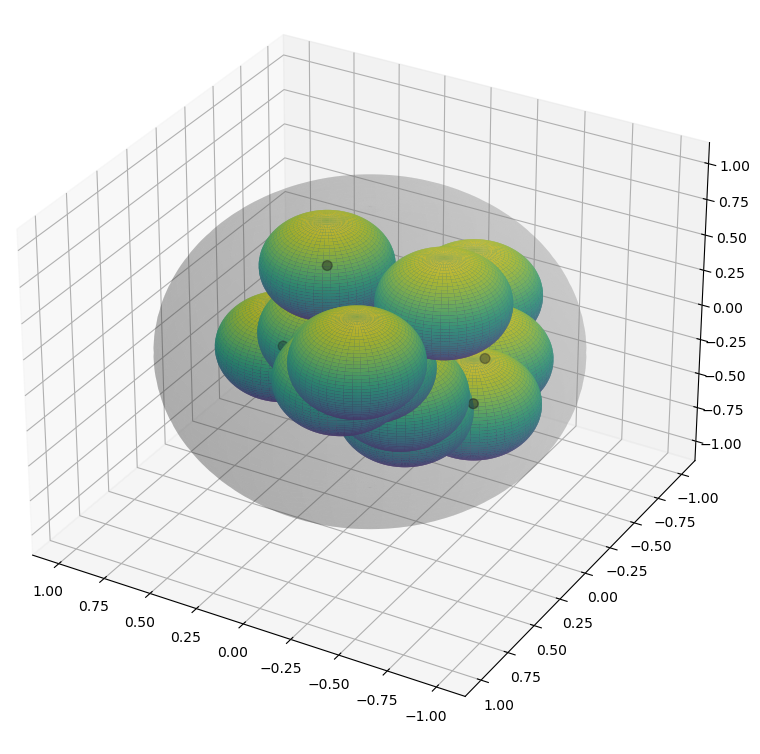}}
\hfill
\subfigure[The final packing of $15$ spheres in the sphere]{\includegraphics[width=0.45\linewidth]{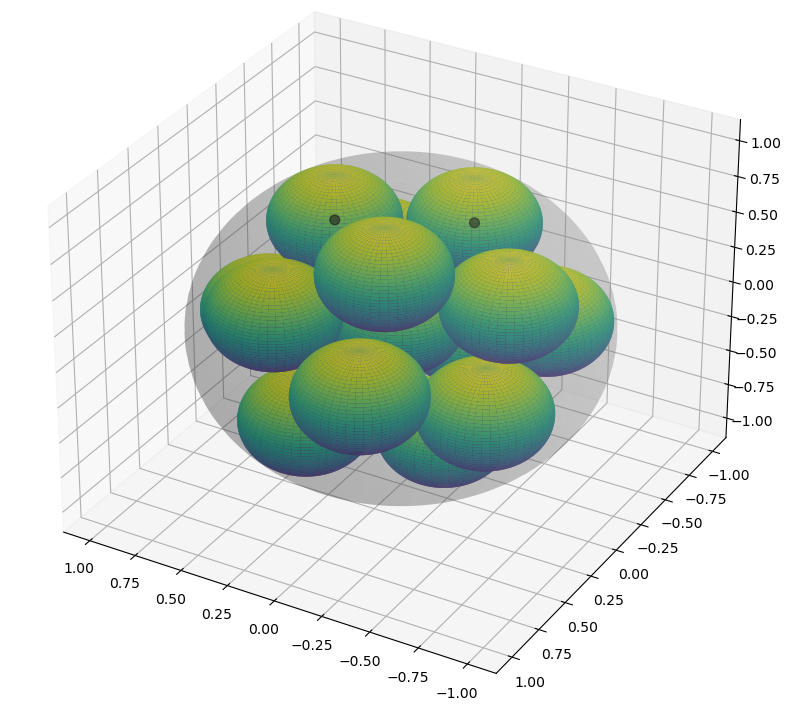}}
\caption{An illustration of initial and final packing of spheres in a sphere and a cube.}
\label{fig:initialfinal}
\end{figure}

Fig.~\ref{fig:initialfinal} illustrates the packing of $15$ spheres of radius, $0.318304823$ units within a sphere of radius $1$ unit, and  that of $15$ spheres of radius, $0.192307692$ units within  a cube of side length $1$ unit. The initial configuration exhibits a high level of overlap among the spheres, influenced by the use of the Xavier uniform initializer for neural network weight initialization. However, as the optimization progresses, the overlap gradually decreases, leading to a notably improved and satisfactory packing configuration.

\newpage 
 \section{Conclusions and Future Work}
In this paper, we have presented a novel encoder-decoder architecture for   packing identical smaller circles within a larger object. Our approach consists of an encoder block that obtains the center of each smaller circles, a perturbation block that adds controlled perturbation to them, and a decoder block that outputs an estimation of the index of the circle being packed. We utilized neural networks to parameterize the encoder and decoder, optimizing them to minimize the error between the estimated and true indices of the packed circles.
By properly tuning the normalization layer at the end of the encoder  and the random variable at the perturbation block, our approach can pack smaller circles  within a larger object of different shapes. 
Experimental results show that our proposed approach provides sub-optimal solutions with competitive performance compared to classical methods for packing circles within a larger circle.  Moreover, we extended our approach to more general packing problems, including packing circles within a square and spheres within a sphere and a cube, in addition to the initial task of packing circles within circles.

Future work involves enhancing the results for packing circles within a circle and a square, packing spheres within a sphere and a cube, and exploring the packing of higher-dimensional spheres in various shapes. Additionally, we intend to investigate the packing of squares, cubes, and other higher-dimensional shapes as smaller objects within larger objects.

\small

\newpage
\bibliographystyle{unsrtnat}
\bibliography{Main_paper.bib}

\end{document}